\documentclass{llncs}
\usepackage[a4paper,margin=1.5in,footskip=0.5in]{geometry}
\usepackage{blindtext}
\usepackage{graphicx}
\usepackage[T1]{fontenc}
\usepackage[utf8]{inputenc}
\usepackage{float}
\usepackage{todonotes}
\begin{document}

\title{Neural Comic Style Transfer: Case Study}

\author{Maciej Pęśko \and Tomasz Trzciński}
\institute{Warsaw University of Technology, Warsaw, Poland,\\
\email{mpesko@mion.elka.pw.edu.pl \\ t.trzcinski@ii.pw.edu.pl},\\
}

\maketitle              

\begin{abstract}
The work by Gatys \textit{et al}. \cite{gatys:} recently showed a neural style algorithm that can produce an image in the style of another image. Some further works introduced various improvements regarding generalization, quality and efficiency, but each of them was mostly focused on styles such as paintings, abstract images or photo-realistic style. In this paper, we present a comparison of how state-of-the-art style transfer methods cope with transferring various comic styles on different images. We select different combinations of Adaptive Instance Normalization \cite{huang:belongie} and Universal Style Transfer \cite{universal:style} models and confront them to find their advantages and disadvantages in terms of qualitative and quantitative analysis. Finally, we present the results of a survey conducted on over 100 people that aims at validating the evaluation results in a real-life application of comic style transfer.

\keywords{Neural Style Transfer, Style Transfer, Comic Style, Comic, Computer Vision, Neural Network}
\end{abstract}

\section{Introduction}
Cartoons and comics became a popular mean of artistic expression worldwide. Unfortunately, only a limited number of talented people with painting or graphics skills can create them holding on to aesthetic standards. What is more, it also takes a significant amount of time to create a valuable comic graphics. Providing an automatic tool to transfer the style of images or videos to comics could revolutionize the way publishers and individuals create {\it comixified} content.

Gatys \textit{et al.} \cite{gatys:} showed that spectacular results in terms of transferring the style between the images can be obtained using convolutional neural networks and their intermediate representations. Further works presented improved versions of the original style transfer algorithm that improve the model in terms of execution speed, results quality and generalization to various styles. Although many publications focused on various applications of style transfer, to our best knowledge, this is the first attempt to evaluate and compare the results obtained by several methods in the context of transferring comic style.

More precisely, in this paper we compare several modern style transfer methods and evaluate their effectiveness in terms of how well they propagate various comic style characteristics between the images. In our work, we focus mostly on the most efficient methods, {\it i.e.} the methods whose execution time per image do not exceed 2 seconds, that enable arbitrary style transfer with infinite number of possible comic styles. 

The remainder of this paper is organized in the following manner. In Section 2 we describe all related works that discuss the topic of style transfer. Section 3 presents an detailed overview of a selected set of the state-of-the-art approaches. Section 4 describes our experimental setup, image collection process and implementation sources. Section 5 presents the results of our evaluation. Section 6 shows results of conducted survey. In Section 7 we make some conclusions and plans for further research.

\section{Related Work}
Gatys \textit{et al.} \cite{gatys:} in their paper demonstrated that deep convolutional neural networks are able to encode the style information of any given image. Moreover, they showed that the content and style of the image can be separated and treated individually. As a consequence, it is possible to transfer the characteristics of the style of a given image to another one, while preserving the content of the latter. They propose to exploit the correlations between the features of deep neural networks in terms of Gram matrices to capture image style. Furthermore, Y. Li \textit{et al.} showed that covariance matrix can be as effective as Gram matrix in representing image style \cite{cov:gram} and theoretically proved that matching the Gram matrices of the neural activations can be seen as minimizing a specific Maximum
Mean Discrepancy \cite{demystif:} function, which gives more intuition on why Gram matrix can represent an artistic style.

Since the original work~\cite{gatys:}, numerous  improvements have been made in the field of Style Transfer. Johnson \textit{et al.} \cite{johnson:} and Ulyanov \textit{et al.} \cite{ulyanov:1} proposed a fast approach that increases the efficiency of style transfer by three orders of magnitude. Nevertheless, this improvement comes at a price of lower results quality. Multiple authors tried to address this shortcoming~ \cite{ulyanov:2,inter:layer,multimodal:,histogram:losses} or make it more generic to enable using different styles in one model. The proposed solution include using a conditional normalization layer that learns normalization parameters for each style \cite{dumoulin:}, swapping a given content feature with the closest style feature in its local neighborhood~\cite{chen:schmidt}, directly adjusting the content feature to match the mean and variance of the style feature \cite{huang:belongie,desai:,ghiasi:} or building a meta network which takes in the style image and produces corresponding image transformation network directly \cite{meta:net}. Other methods include an automatic segmentation of the objects and extraction of their soft semantic masks \cite{zao:rosik} or adjusting feature maps using whitening and coloring transforms \cite{universal:style}. There are also some works that try to make photo-realistic style transfer \cite{photo:real,deep:photo} to carry style of one photo to another, leaving it as realistic as possible. In addition, many works have been created that focus on some other, various fields of Style Transfer. For instance Coherent Online Video Style Transfer \cite{chen2017coherent} which describes end-to-end network that generates consistent stylized video sequences in near real time, StyleBank \cite{chen2017stylebank} which uses multiple convolution filter banks, where each filter in bank explicitly represents one style or Stereoscopic Neural Style Transfer \cite{chen2018stereoscopic} that concerns on 3D or AR/VR subject. 

It is worth mentioning that existing approaches often suffer from a trade-off between generalization, quality and efficiency. More precisely, the optimization-based approaches handle arbitrary styles with a great visual quality, but the computational costs are relatively high. On the other side, the feed-forward methods are executed very fast with slightly worse but acceptable quality, but they are limited to a specific, fixed number of styles. Finally, the arbitrary methods are fast and enable multiple style transfer, but often their quality is worse than the previously proposed ones.
\section{Methods Overview}

In this section, we give an overview of the methods proposed for transferring the style of images. We start with the initial work of Gatys {\it et al.}~\cite{gatys:} and continue with the follow-up works.

\subsection{Style transfer}
Gatys \textit{et al.} \cite{gatys:} in their work, use 16 convolutional and 5 pooling layers of the 19-layer VGG network \cite{vgg:}. They pass three images through this network: Content image, Style image and White Noise image. The content information is given by feature maps from one layer 
and content loss is the squared-error loss between the two feature representations:
\begin{equation}
  L_{content} = \frac{1}{2} \sum_{i,j} (F_{ij}^{l} - P_{ij}^{l})^2 ,
\end{equation}

Where $F_{ij}^{l}$ is the activation of the i-th filter at position $j$ in layer $l$. On the other side, the style information is given by the Gram matrix $G$ of the vectorized feature maps $i$ and $j$ in layer $l$:
\begin{equation}
  G_{ij}^{l} = \sum_{k} F_{ik}^{l} F_{jk}^{l} ,
\end{equation}

The style loss is a squared-error loss function computed between two Gram matrices obtained from specific layers from white noise image and style image passed through the network. $N$ is a number of feature maps and $M$ is a feature map size.
\begin{equation}
  L_{style} = \frac{1}{2} \sum_{l=0}^L \omega_lE_l ,
\end{equation}
\begin{equation}
  E_{l} = \frac{1}{4N_l^2M_l^2} \sum_{i,j} (G_{ij}^{l} - A_{ij}^{l})^2 ,
\end{equation}

Finally, the cost function is defined as weighted sum of two above losses. Namely, between the activations (content) and Gram matrices (style) and then is minimized using backpropagation.
\begin{equation}
  L_{total} = \alpha*L_{content} + \beta*L_{style}
\end{equation}

Unfortunately, this method is very slow. Using this approach, style transfer of one 512x512 pixel image lasts almost a minute on recent GPU architectures, such as NVIDIA Quadro or Titan X~\cite{johnson:,demystif:}. It is caused by the fact that for each pair of images, this method performs an optimization process using backpropagation to minimize $L_{total}$. To address this shortcoming, several approaches have been proposed, including Adaptive Instance Normalization, described below.
%
\subsection{Adaptive Instance Normalization}
X. Huang and S. Belongie in their paper \cite{huang:belongie} present the first fast and arbitrary neural style transfer algorithm that resolves problem with generalization, quality and efficiency trade-off. This method consists of two networks: a style transfer network and a loss network. 

\begin{figure}[ht!]
  \centering
  \includegraphics[width=0.9\textwidth]{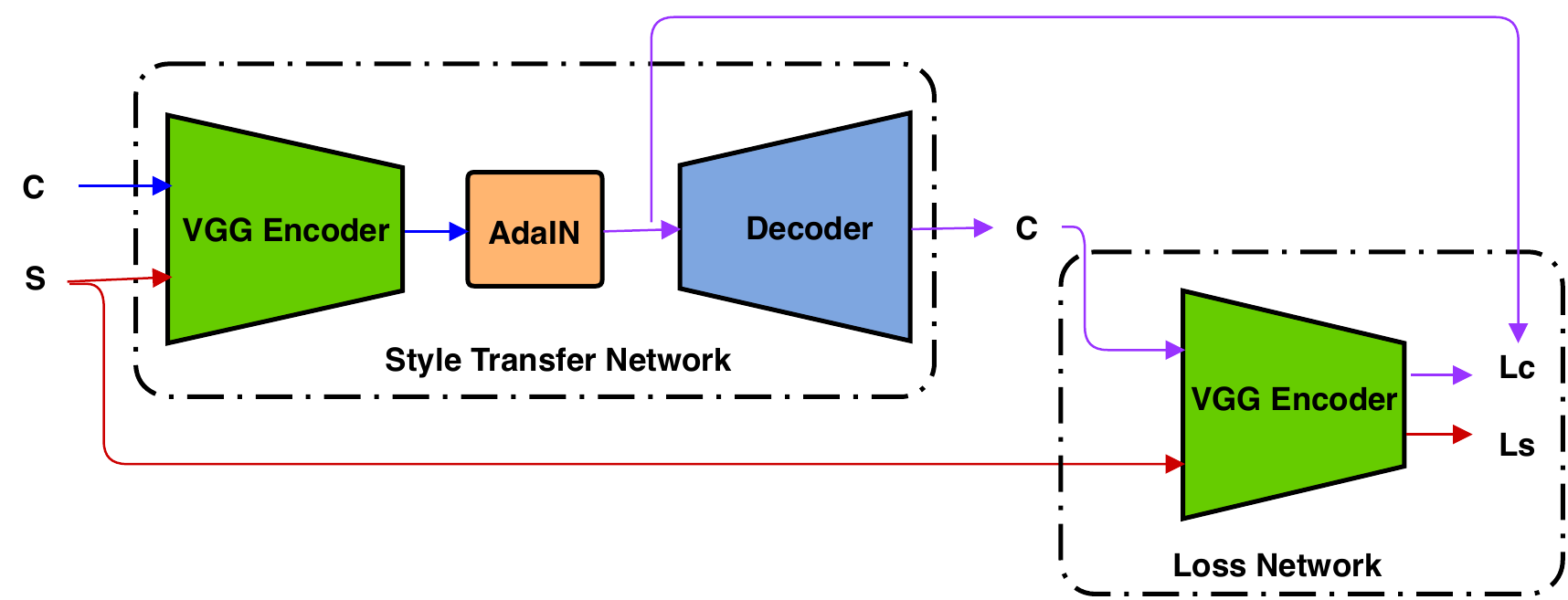}
  \caption{AdaIN architecture. C and S are Content and Style images and Lc and Ls are content and style losses. \label{AdaIN:arch}}
\end{figure}

The loss network is pretty similar to the network presented in~\cite{gatys:}. It is used to compute a total loss which is minimized by a backpropagation algorithm to fit the parameters of a style transfer network. The style transfer network consists of a simple encoder-decoder architecture. The encoder is first few layers of a pre-trained VGG-19 network. The decoder mirrors the encoder with a few minor differences. All pooling layers are replaced by nearest up-sampling layers and there are no normalization layers. The most interesting part is AdaIN layer which is placed between the encoder and the decoder. AdaIN produces the target feature maps that are inputs to the decoder by aligning the mean and variance of the content feature maps to style feature maps. A randomly initialized decoder is trained using a loss network to map outputs of AdaIN to the image space in order to get a stylized image. AdaIN architecture can be seen in the Fig. \ref{AdaIN:arch}. However, this method only aligns the most basic feature statistics which are mean and variance. It is only approximation of the Gram matrix approach presented in ~\cite{gatys:}. Fortunately, subsequent works have introduced some changes in this matter.

\subsection{Universal Style Transfer}
\begin{figure}[H]
  \centering
  \includegraphics[width=1.0\textwidth]{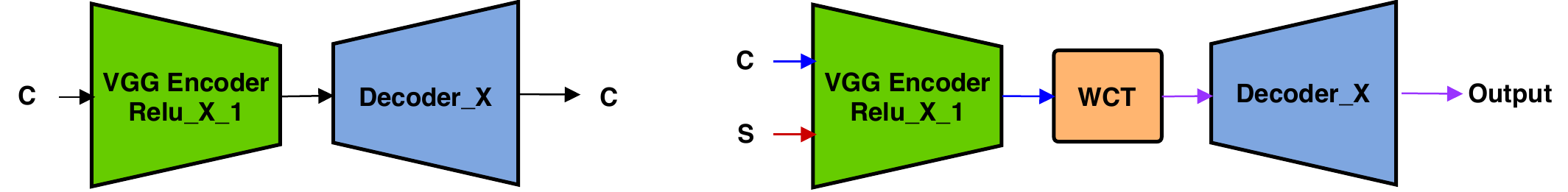}
  \caption{Training process (left hand side) and single level encoder-WCT-decoder network (right hand side). C and S are content and style images, respectively. \label{ust:arch1}}
\end{figure}

Knowing that covariance matrix can be as effective as Gram matrix in representing image style \cite{cov:gram}, Y. Li \textit{et al.} \cite{universal:style} proposed a novel approach called Universal Style Transfer (UST). It is closely related to AdaIN method but the intuition of UST is to match covariance matrices of feature maps instead of aligning only mean and variance which was proposed in \cite{huang:belongie}. Their Universal Style Transfer approach (UST) uses a very similar encoder-decoder architecture where encoder is composed of a first few layers of a pre-trained VGG-19 network and the decoder mostly mirrors the encoder. However, instead of using the AdaIN layer to carry style information, they use Whitening and Coloring transform (WCT). Moreover, in this method style transfer that is represented by WCT layer is not used during training. The network is only trained to reconstruct an input content image. The entire style transfer takes place in the WCT layer which is added to already pre-trained image reconstruction network.f Training process and a single encoder-decoder network is presented in Fig.~\ref{ust:arch1}.

The Universal Style Transfer can be described in the following way. Firstly, five different encoder-decoder networks are pre-trained for image reconstruction purpose. Each of them consists of a different number of VGG-19 layers in encoder and the same number in decoder 
Afterwards, WCT is placed as an intermediate element between the encoder and the decoder. Given a pair of style image $I_s$ and content image $I_c$, encoder extracts their vectorized feature maps $f_c$ and $f_s$, for content and style images, respectively. Next, the WCT module is used to adjust $f_c$ to match the statistics of $f_s$ or more precisely, directly transform the $f_c$ to match the covariance matrix of $f_s$. Finally, the decoder reconstructs the image using transformed feature maps.

\begin{figure}[H]
  \centering
  \includegraphics[width=0.7\textwidth]{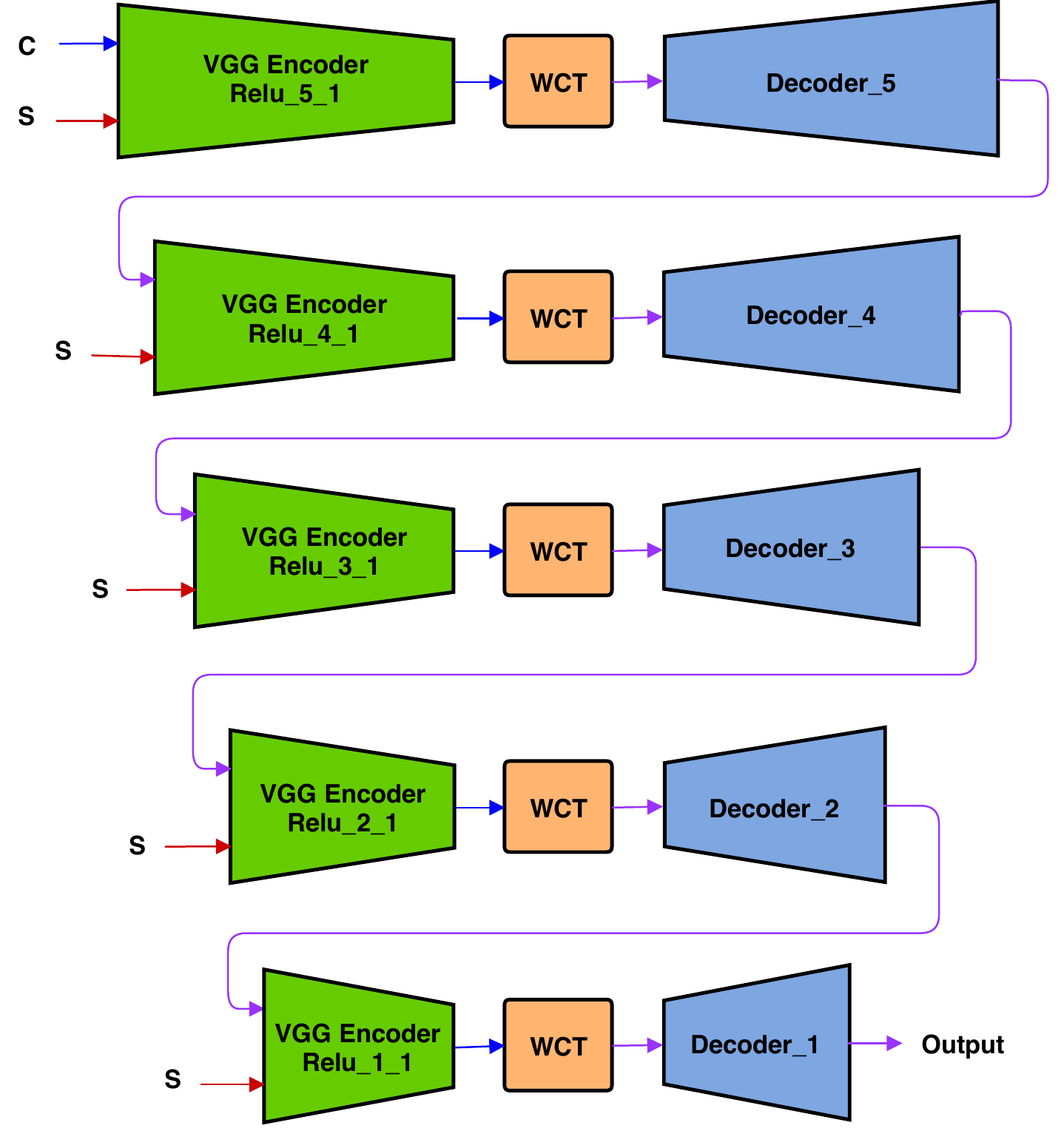}
  \caption{Universal Style Transfer architecture with the whole multi level pipeline. Each level of stylization consists of single encoder-WCT-decoder network with different decreasing number of VGG layers. C and S are content and style images, respectively. \label{ust:arch2}}
\end{figure}

Another extension is to use a multi-level stylization in order to match the statistics of the style at all abstraction levels. It means that the result obtained from a network that matches higher level information is treated as the new content input image to network that matches lower level statistics. In the original paper \cite{universal:style}, UST flow consists of 5 level stylization where stylization is a one single pass through the encoder-WCT-decoder network. Such architecture can be seen in Fig. \ref{ust:arch2}. In our work, except for the 5-level architecture, we also use a shallower version with 4 levels in order to try this method with less style abstraction. What it means is to not use first step (Relu\_5\_1 - WCT - Decoder\_5 in Fig. \ref{ust:arch2}) of 5-level architecture that is responsible for transferring the highest level statistics of the style. We believe that using this modification we can get less distortions in output image. We also use the UST approach with AdaIN layer instead of WCT.

\bigskip
\textbf{Whitening and Coloring Transform.}
Firstly, all vectorized feature maps $f_c$ are centered by subtracting its mean vector. Then $f_c$ are linearly transformed using Whitening Transform so that we obtain $\widehat{f}_c$ in such a way that the feature maps $\widehat{f}_c$ are uncorrelated and each of them has a variance equal to 1.0. Afterwards, all vectorized style feature maps $f_s$ are centered by subtracting its mean vector $m_s$. Subsequently, Coloring Transform is performed. Essentially, it is the inverse of the Whitening step that is used to transform $f_c$ to $\widehat{f}_c$. $\widehat{f}_c$ are transformed to a vector $\widehat{f}_{cs}$ with a specified covariance matrix corresponding to vectorized style feature maps $f_s$. Finally, the $\widehat{f}_{cs}$ is re-centered with the mean vector $m_s$ and the output is a vector $\widehat{f}_{cs}$ which is a transformed vector $f_c$ to match the covariance matrix of $f_s$. 

\subsection{Photorealistic Image Stylization}
\begin{figure}[H]
  \centering
  \includegraphics[width=0.8\textwidth]{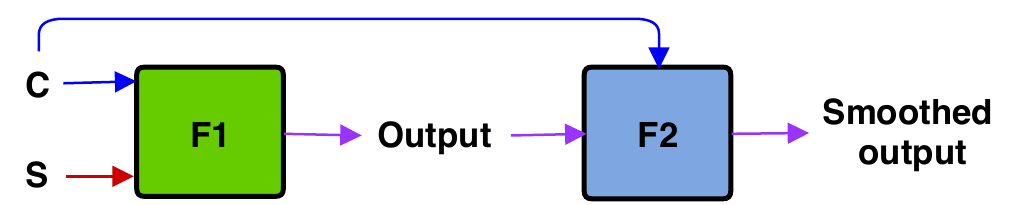}
  \caption{PHOTO-R architecture. $I_s$ and $I_c$ are style and content images. \label{photo-r-arch}}
\end{figure}

In our paper, we would like to evaluate different types of neural style transfer models in the field of comic style transfer. Therefore, we also evaluate a Photorealistic Image Stylization (PHOTO-R) method proposed by Y. Li \textit{et al.} in their paper \cite{photo:real}. Their method is strongly connected with UST with some minor differences. However, those slight changes cause their model to give more photorealistic output images. The PHOTO-R algorithm consists of two steps that can be seen in Fig. \ref{photo-r-arch}. The first step (F1) is a modified version of UST with four levels of stylization. Less stylization layers means that less abstraction information is transfered so more details of the input image are preserved. Moreover, F1 differs from UST also by using unpooling layers instead of upsampling layers. This change is crucial to the results obtained by this model and causes the model to preserve much more spatial information. The second step (F2) is a smoothing process that removes visible artifacts and returns a photorealistic image. F2 smooths pixels in a local neighborhood and ensures that if those pixels have similar content then they are stylized in a similar way. 

\section{Experimental Setup}
For our comparison we choose five state-of-the-art methods, discussed in previous chapter. Each of them performs fast and arbitrary style transfer. The selected methods are:
\begin{itemize}
\item Adaptive Instance Normalization (AdaIN) \cite{huang:belongie}
\item Universal Style Transfer (UST-WCT) \cite{universal:style}
\item Shorten Universal Style Transfer with four levels of stylization instead of five (UST-WCT4)
\item Universal Style Transfer with Adaptive Instance Normalization layer \cite{huang:belongie} instead of Whitening and Coloring Transform \cite{universal:style} (UST-AdaIN)
\item Photorealistic Image Stylization (PHOTO-R) \cite{photo:real}
\end{itemize}

We use pretrained models, provided by authors of the above-mentioned methods. For AdaIN\footnote{\url{https://github.com/xunhuang1995/AdaIN-style}}, UST-AdaIN and UST-WCT\footnote{\url{https://github.com/Yijunmaverick/UniversalStyleTransfer}}, and PHOTO-R\footnote{\url{https://github.com/NVIDIA/FastPhotoStyle}}, we use authors' implementations and for UST-WCT4 we use the Tensorflow implementation\footnote{\url{https://github.com/eridgd/WCT-TF}} with minor modification covering 4-level instead of 5-level architecture.

We set the parameter $\alpha$ controlling style-content transfer as indicated in the original works: 1.0 for AdaIN and UST-AdaIN and 0.6 for UST-WCT and UST-WCT4. We prepare 20 different style and content images collected from the Internet. We select style images to represent different comic styles. Content pictures are chosen randomly from the results of image search on Google. All images are rescaled to size 600x450 pixels and evaluated with all five models using 12GB TITAN X GPU.
\section{Results}
\subsection{Qualitative Results}

\begin{figure}[!ht]
   \centering
\begin{tabular}{ccccccc}

\includegraphics[width=1.9cm]{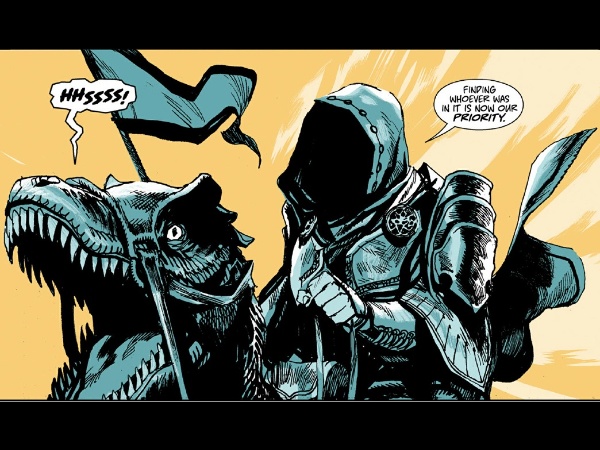}&
\includegraphics[width=1.9cm]{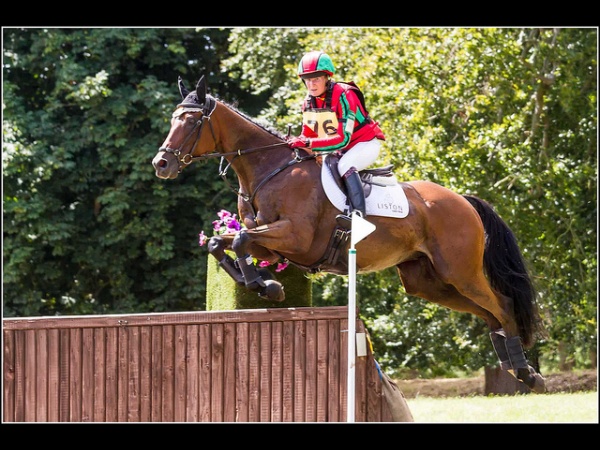}&
\includegraphics[width=1.9cm]{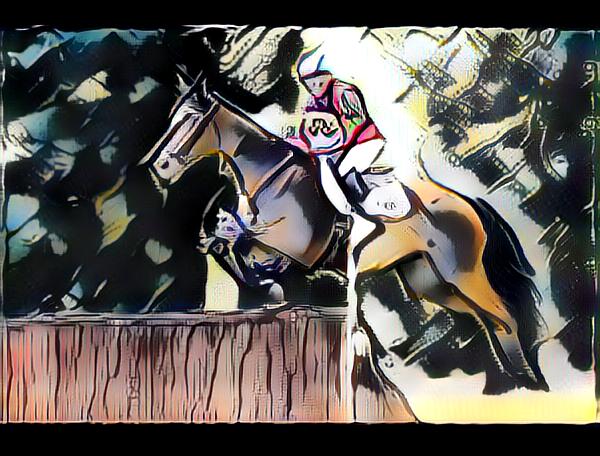}&
\includegraphics[width=1.9cm]{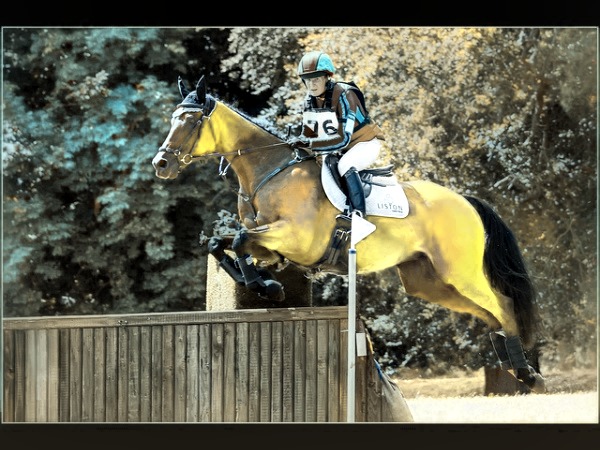}&
\includegraphics[width=1.9cm]{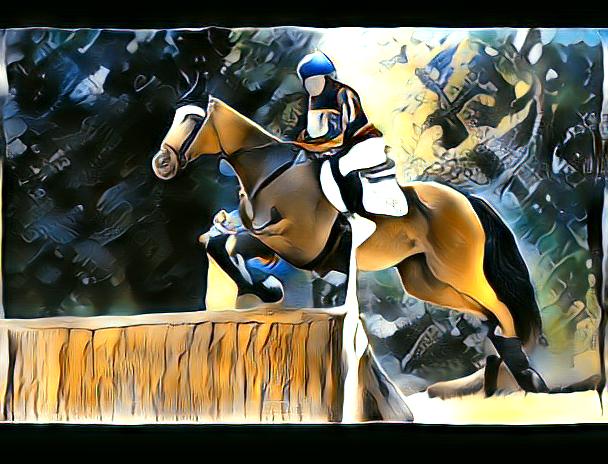}&
\includegraphics[width=1.9cm]{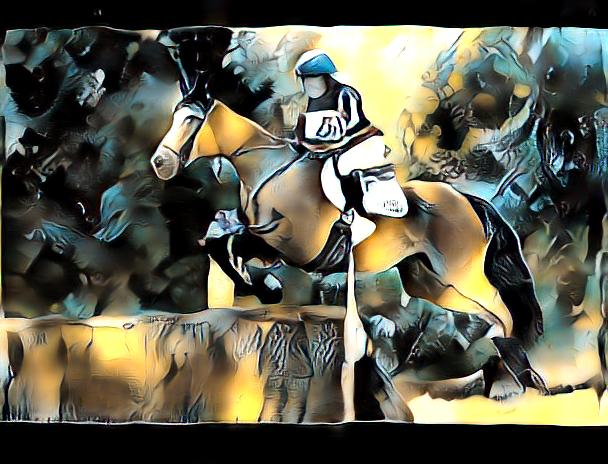}&
\includegraphics[width=1.9cm]{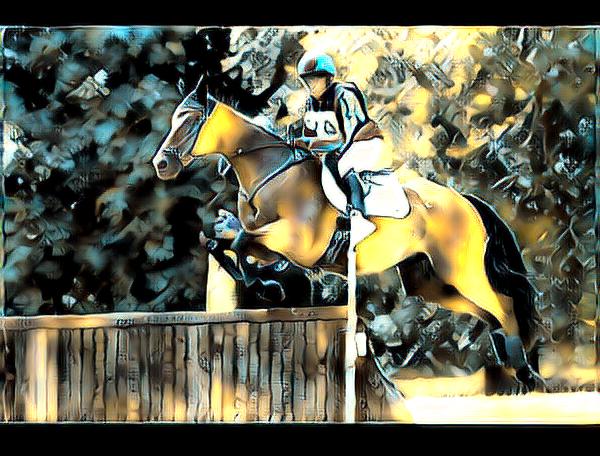}\\

\includegraphics[width=1.9cm]{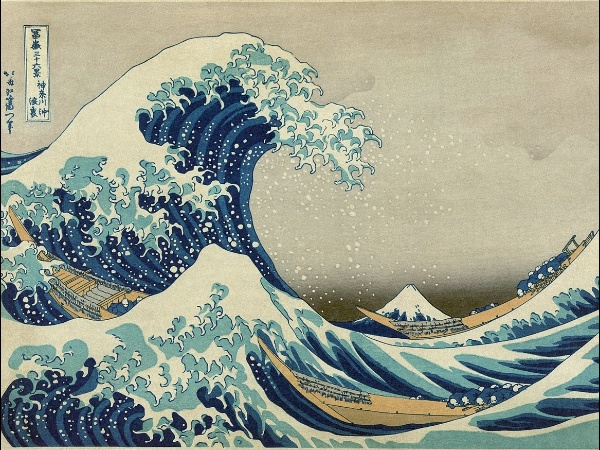}&
\includegraphics[width=1.9cm]{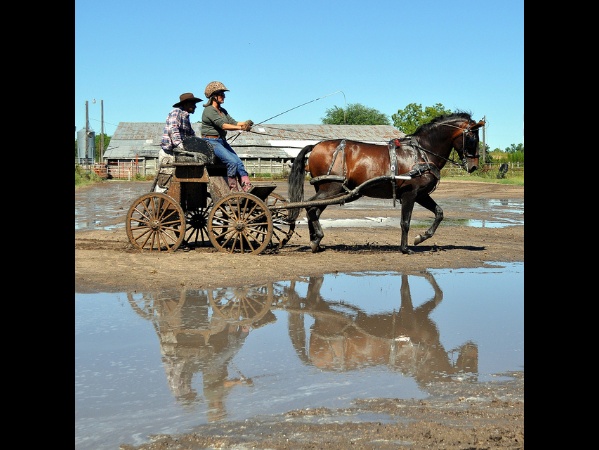}&
\includegraphics[width=1.9cm]{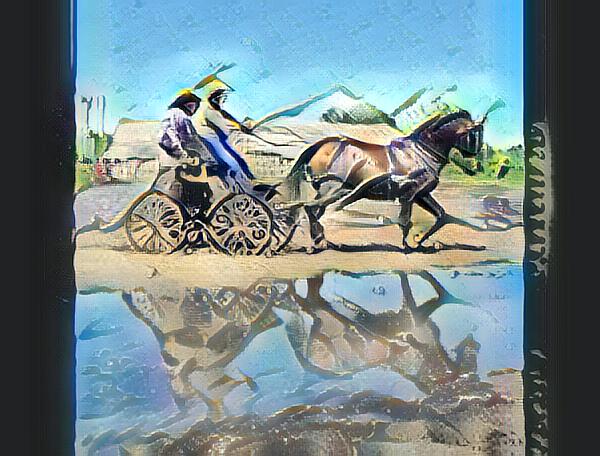}&
\includegraphics[width=1.9cm]{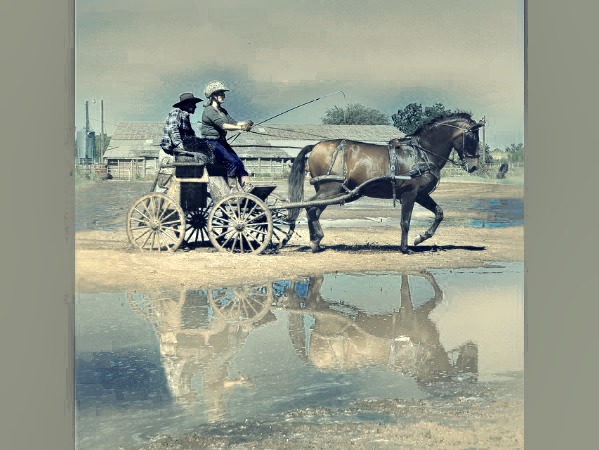}&
\includegraphics[width=1.9cm]{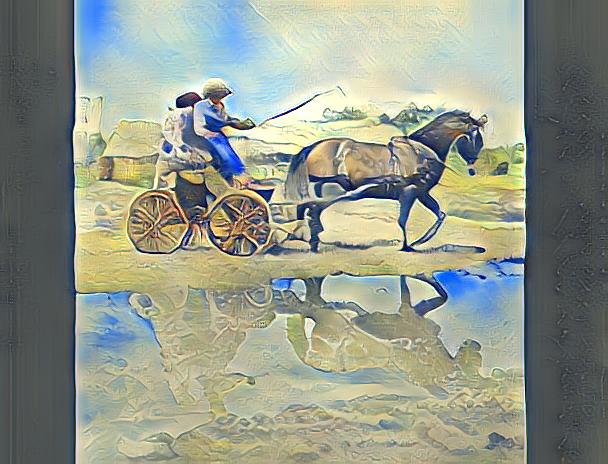}&
\includegraphics[width=1.9cm]{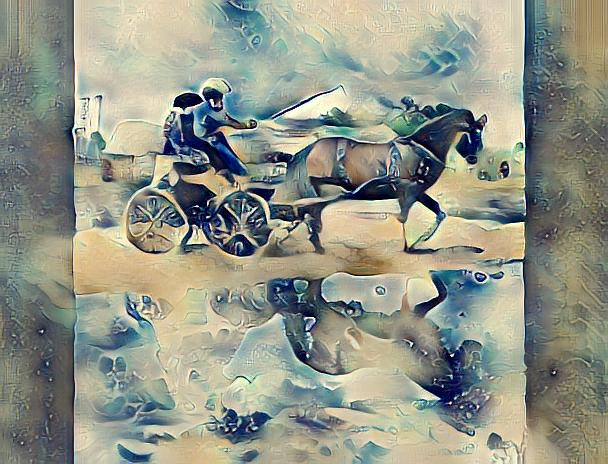}&
\includegraphics[width=1.9cm]{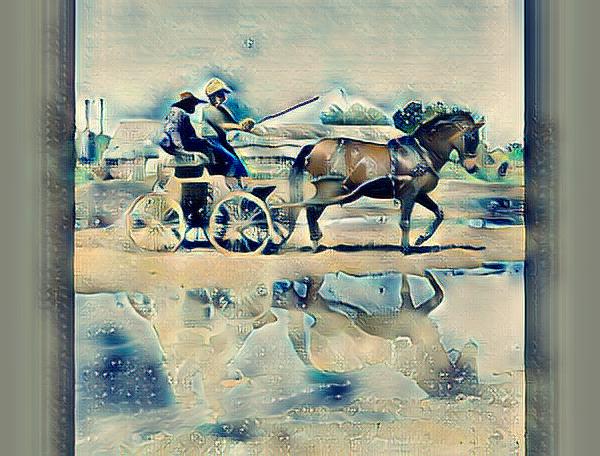}\\

\includegraphics[width=1.9cm]{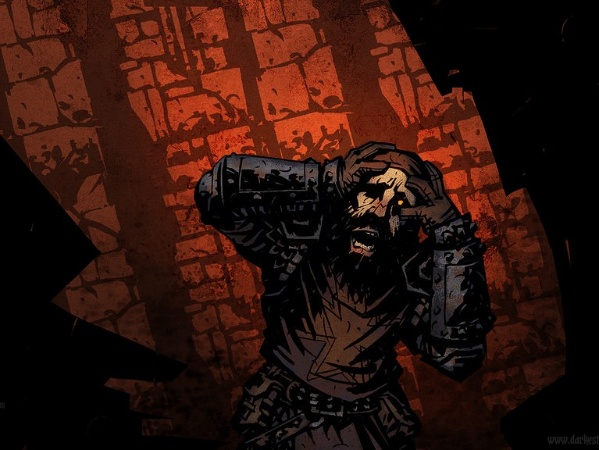}&
\includegraphics[width=1.9cm]{Content/3.jpg}&
\includegraphics[width=1.9cm]{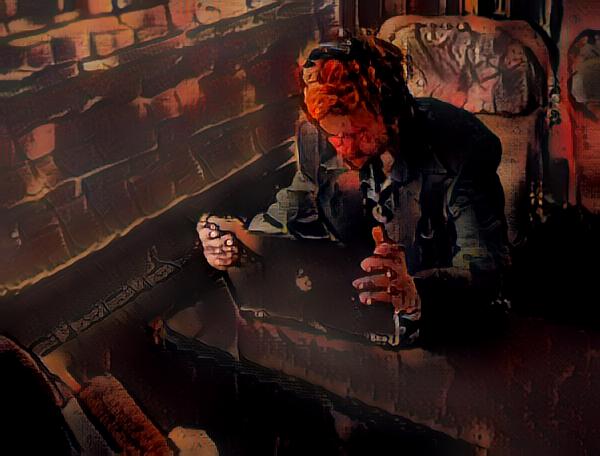}&
\includegraphics[width=1.9cm]{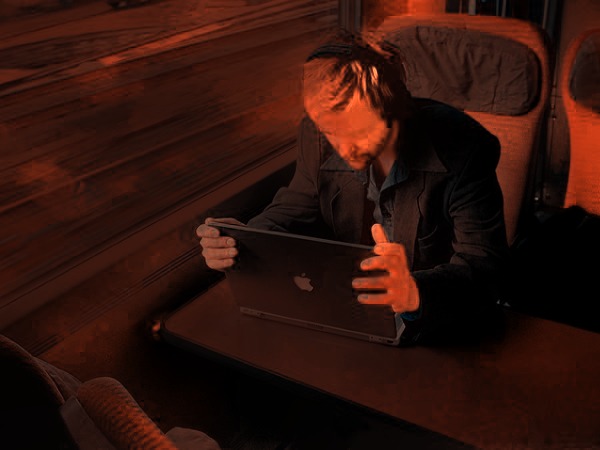}&
\includegraphics[width=1.9cm]{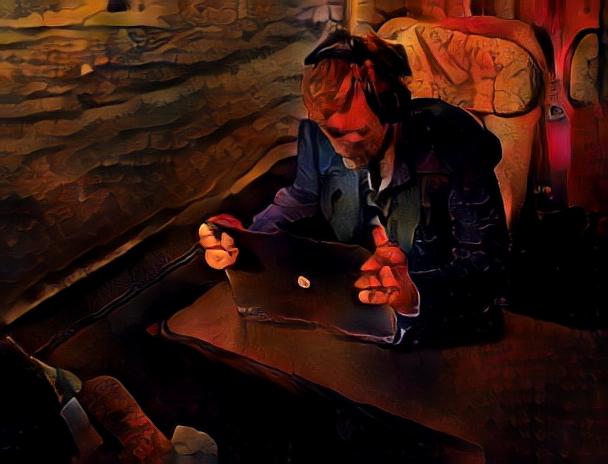}&
\includegraphics[width=1.9cm]{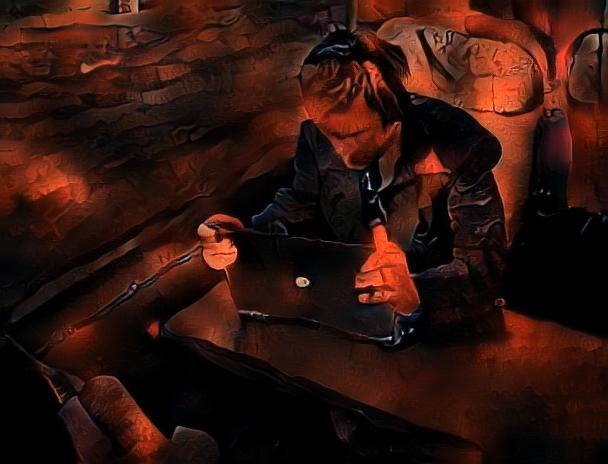}&
\includegraphics[width=1.9cm]{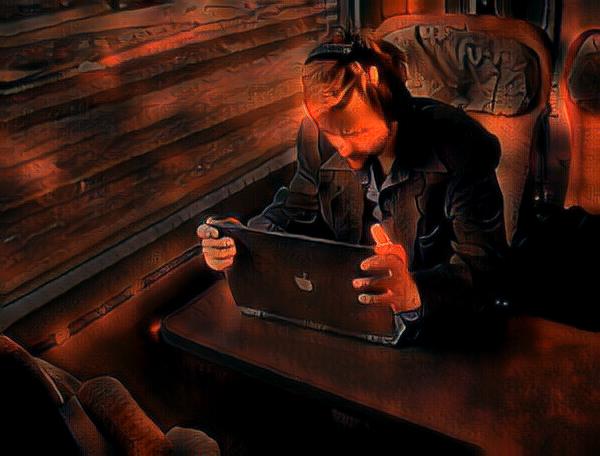}\\

\includegraphics[width=1.9cm]{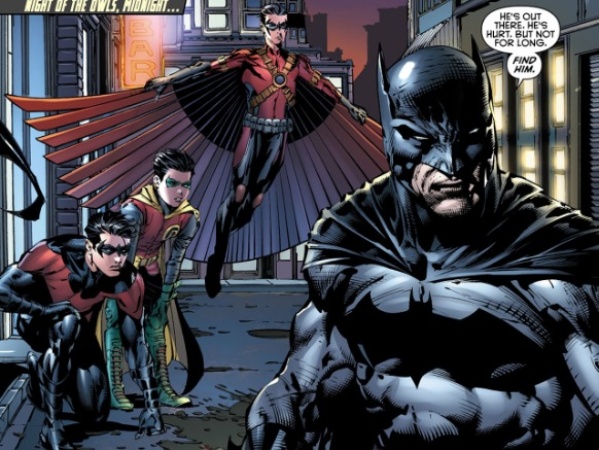}&
\includegraphics[width=1.9cm]{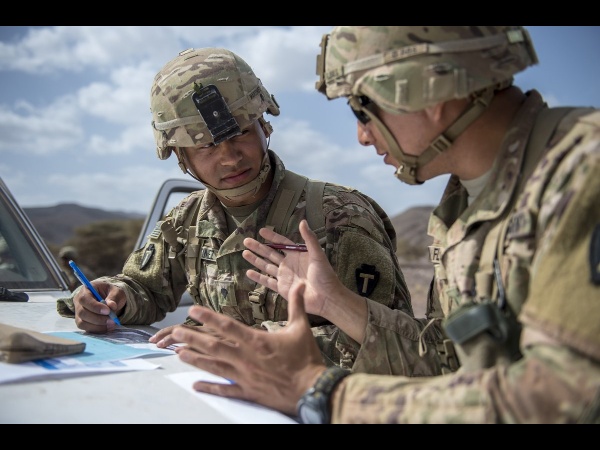}&
\includegraphics[width=1.9cm]{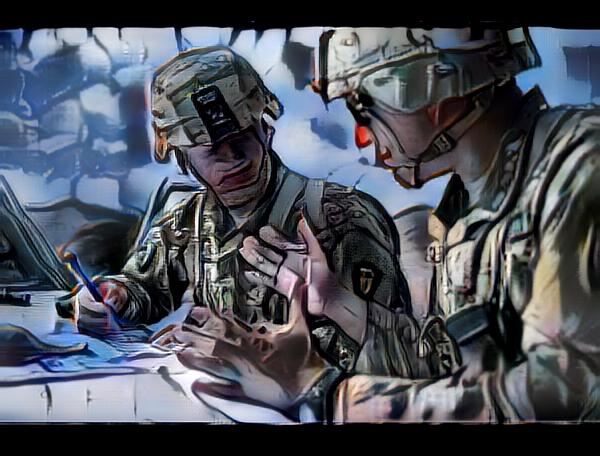}&
\includegraphics[width=1.9cm]{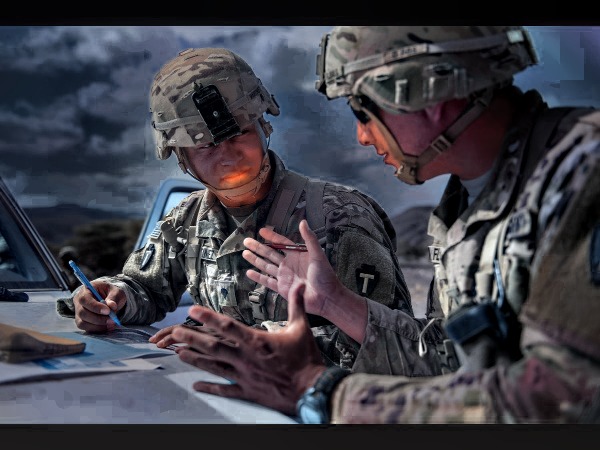}&
\includegraphics[width=1.9cm]{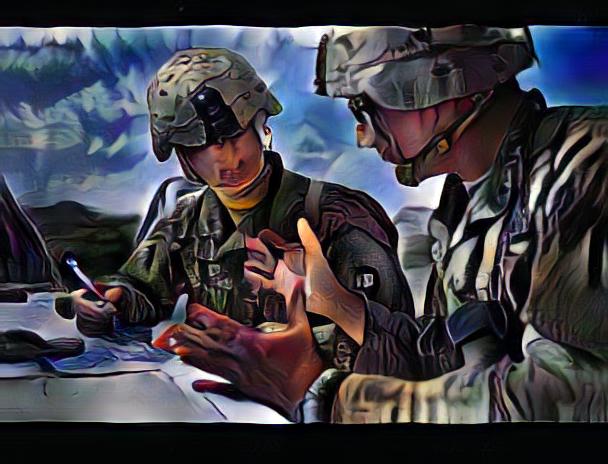}&
\includegraphics[width=1.9cm]{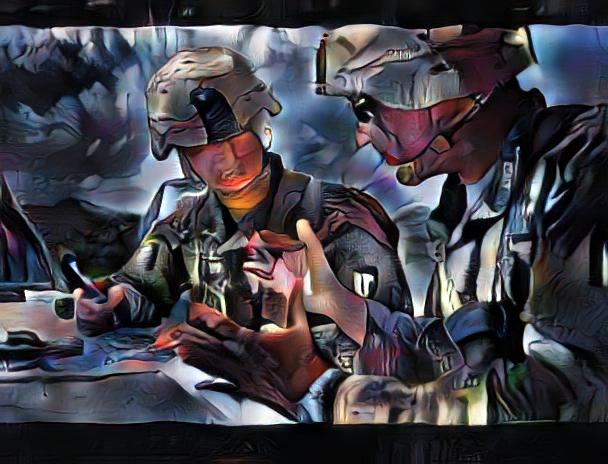}&
\includegraphics[width=1.9cm]{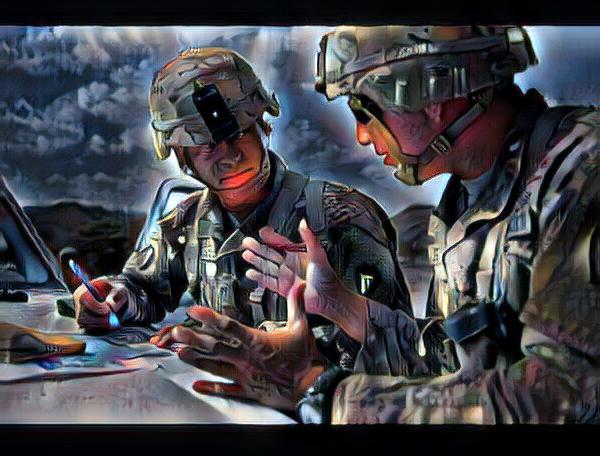}\\

\includegraphics[width=1.9cm]{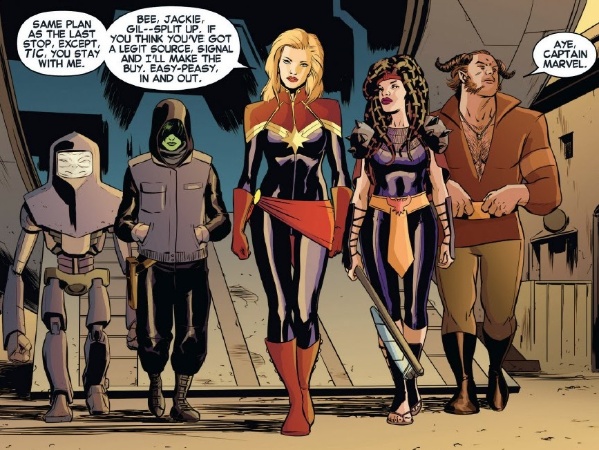}&
\includegraphics[width=1.9cm]{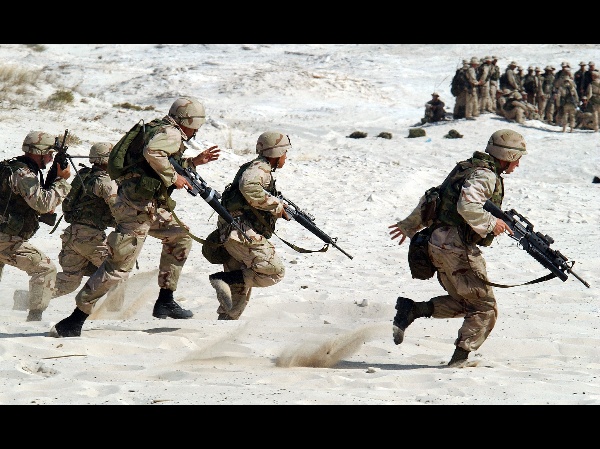}&
\includegraphics[width=1.9cm]{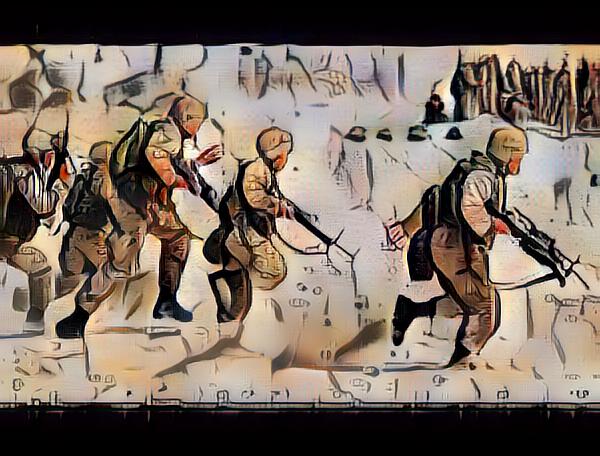}&
\includegraphics[width=1.9cm]{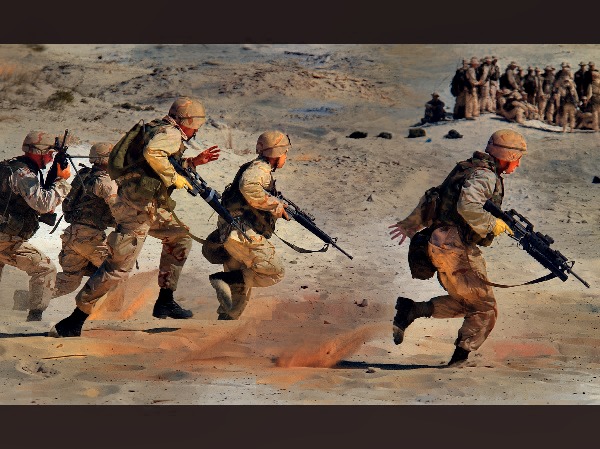}&
\includegraphics[width=1.9cm]{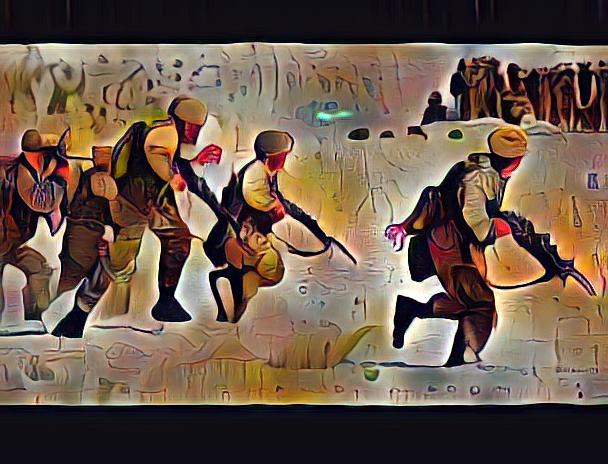}&
\includegraphics[width=1.9cm]{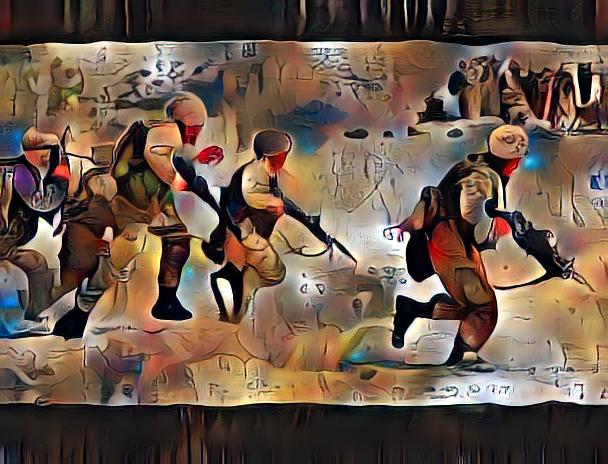}&
\includegraphics[width=1.9cm]{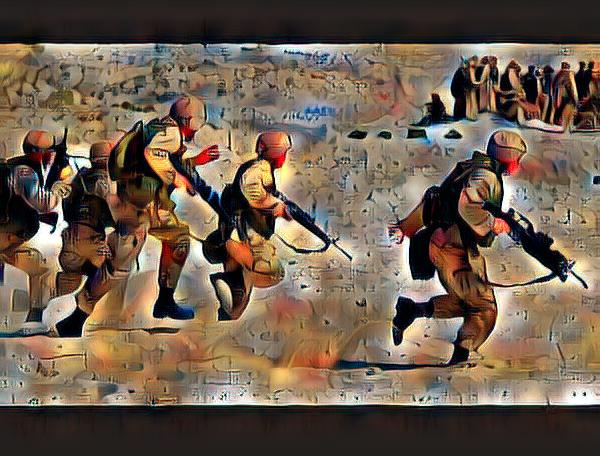}\\

\includegraphics[width=1.9cm]{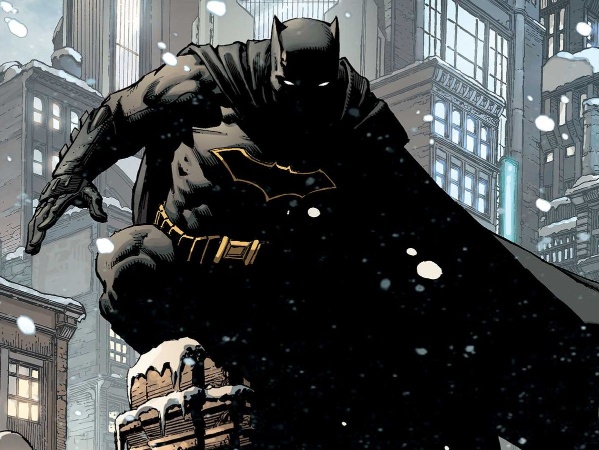}&
\includegraphics[width=1.9cm]{Content/6.jpg}&
\includegraphics[width=1.9cm]{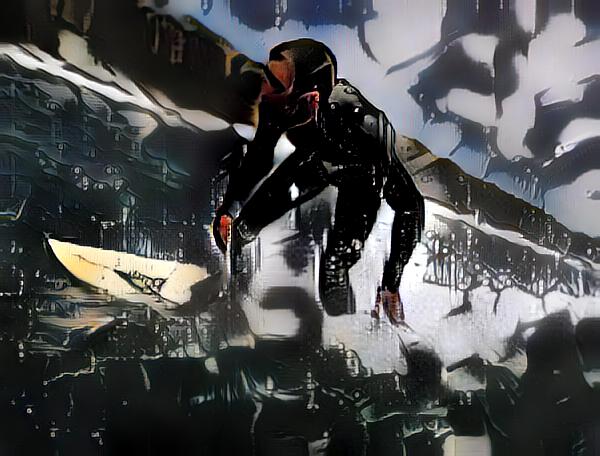}&
\includegraphics[width=1.9cm]{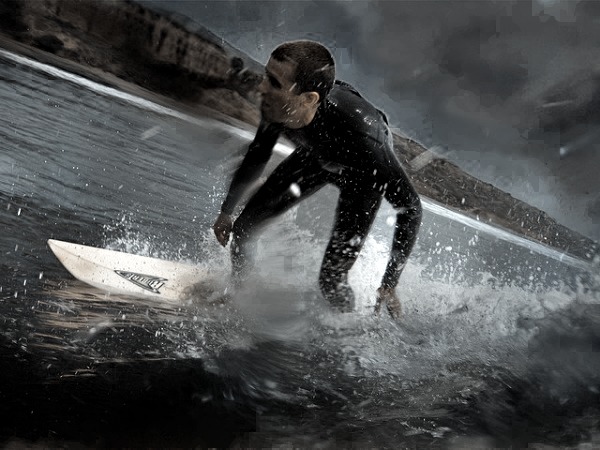}&
\includegraphics[width=1.9cm]{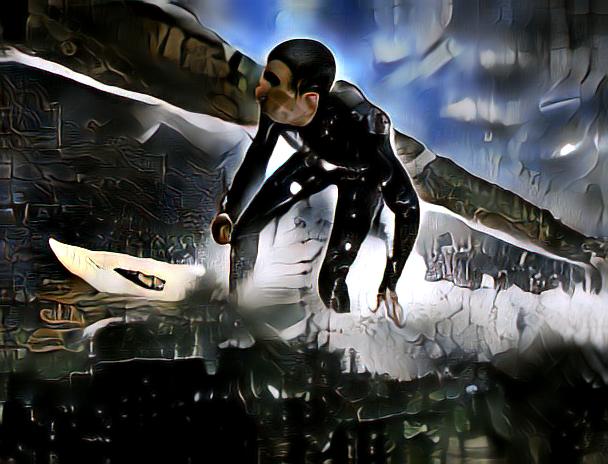}&
\includegraphics[width=1.9cm]{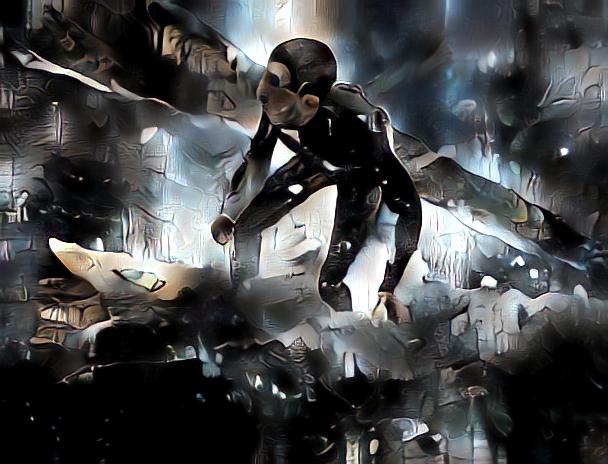}&
\includegraphics[width=1.9cm]{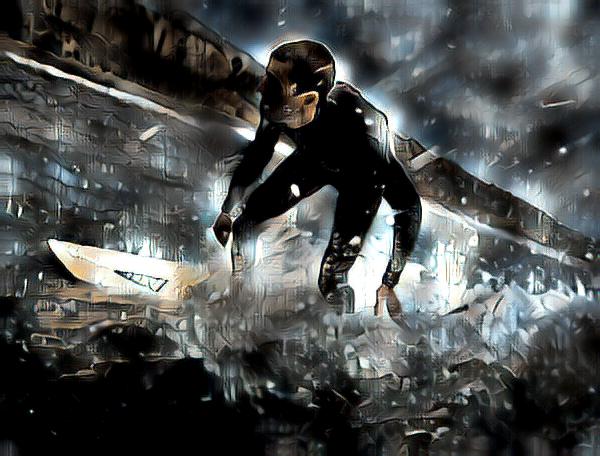}\\

\includegraphics[width=1.9cm]{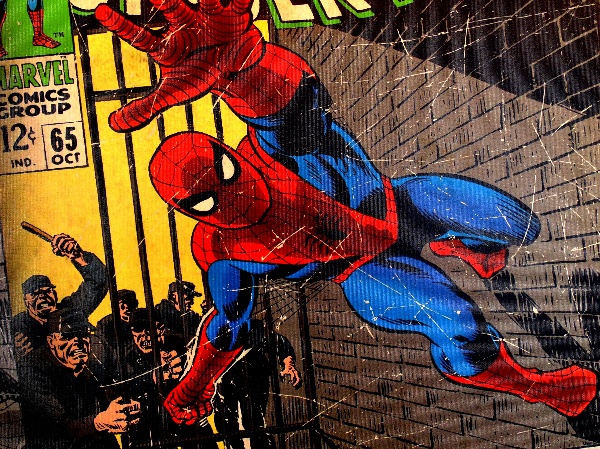}&
\includegraphics[width=1.9cm]{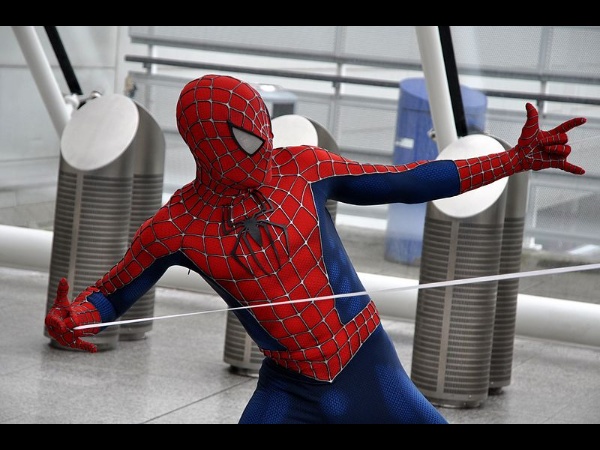}&
\includegraphics[width=1.9cm]{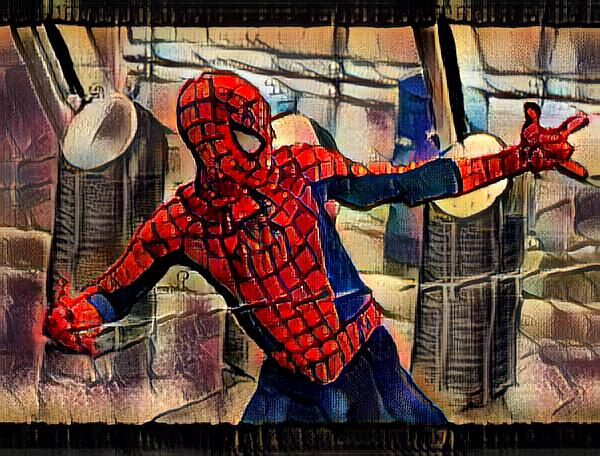}&
\includegraphics[width=1.9cm]{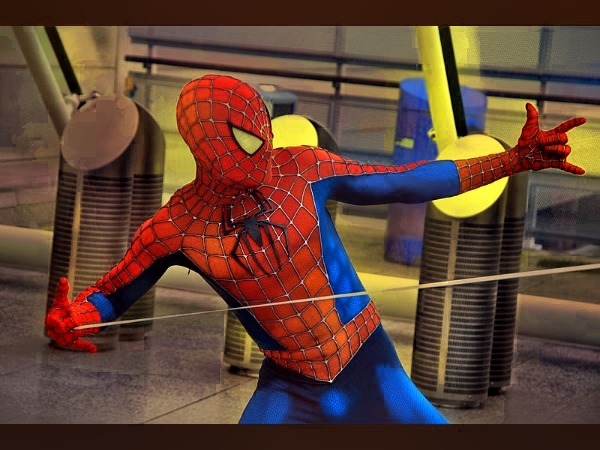}&
\includegraphics[width=1.9cm]{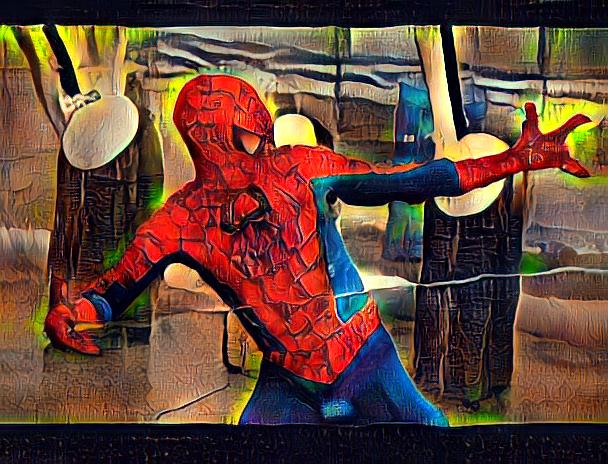}&
\includegraphics[width=1.9cm]{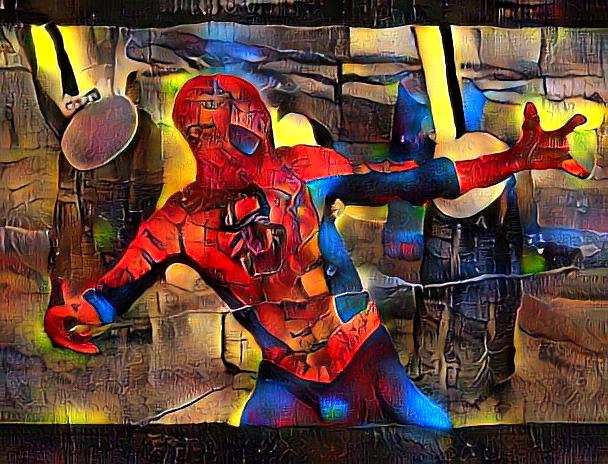}&
\includegraphics[width=1.9cm]{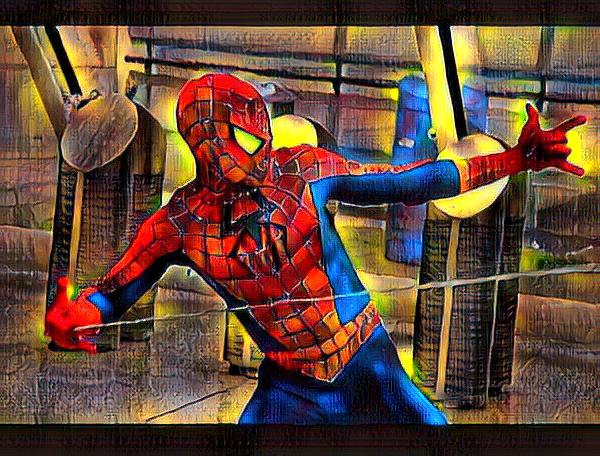}\\

\includegraphics[width=1.9cm]{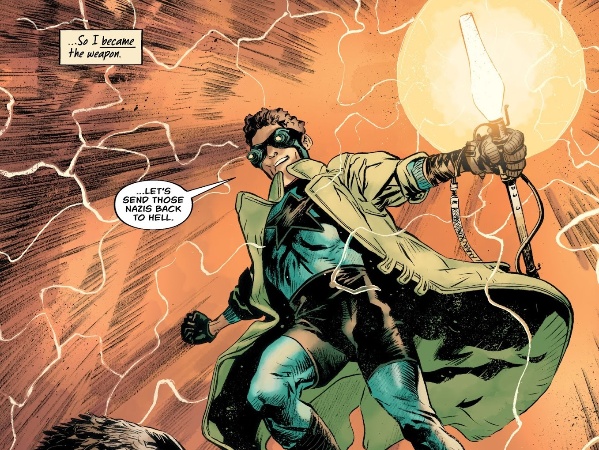}&
\includegraphics[width=1.9cm]{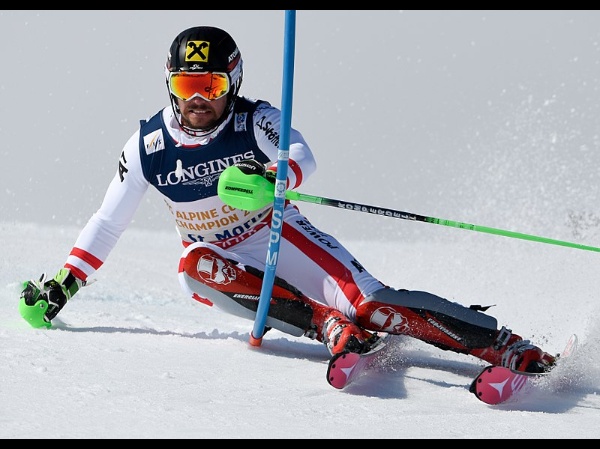}&
\includegraphics[width=1.9cm]{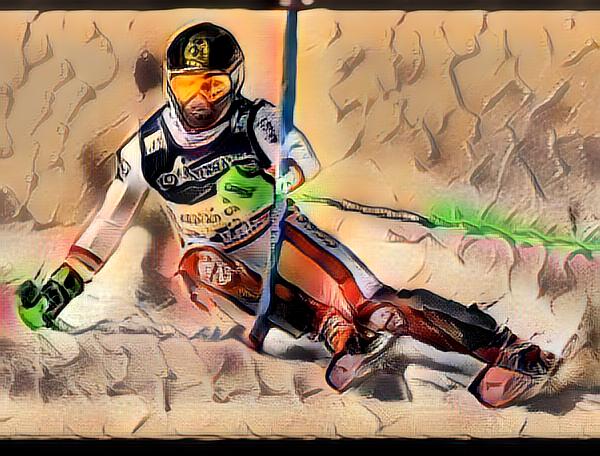}&
\includegraphics[width=1.9cm]{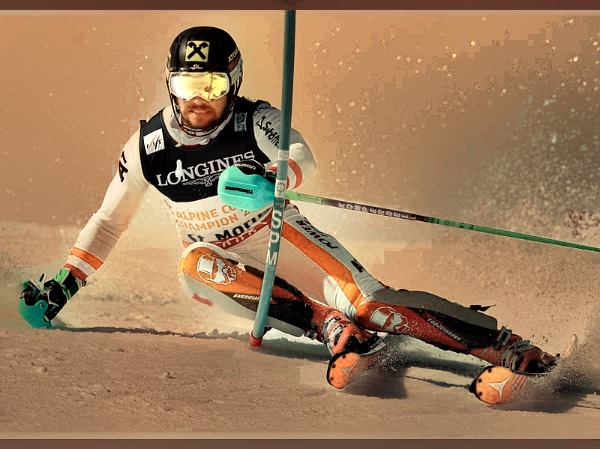}&
\includegraphics[width=1.9cm]{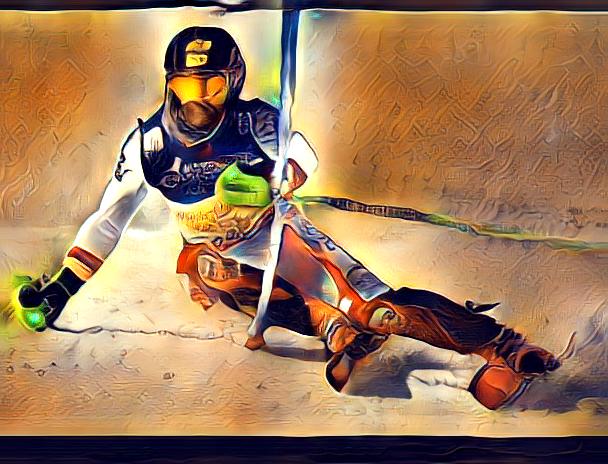}&
\includegraphics[width=1.9cm]{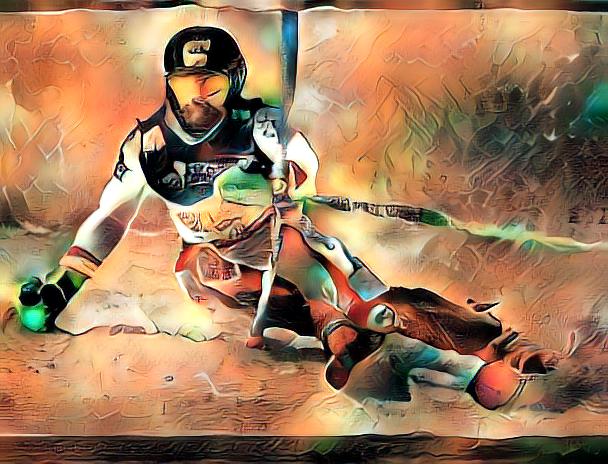}&
\includegraphics[width=1.9cm]{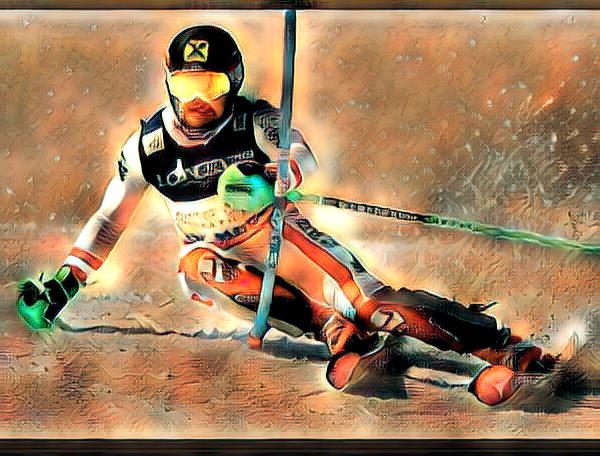}\\

\includegraphics[width=1.9cm]{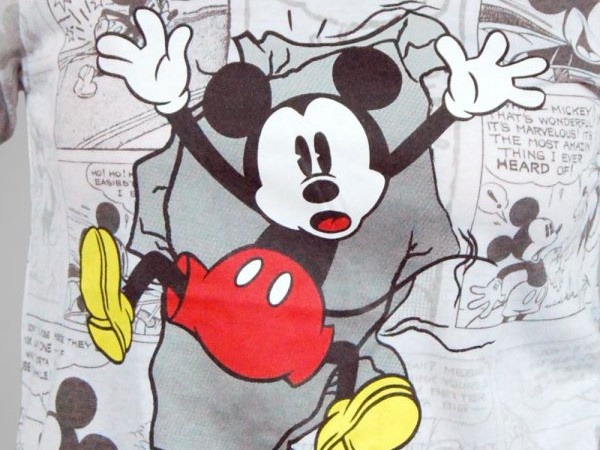}&
\includegraphics[width=1.9cm]{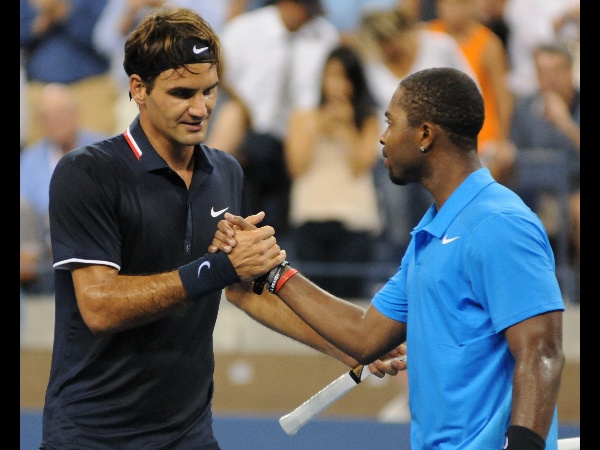}&
\includegraphics[width=1.9cm]{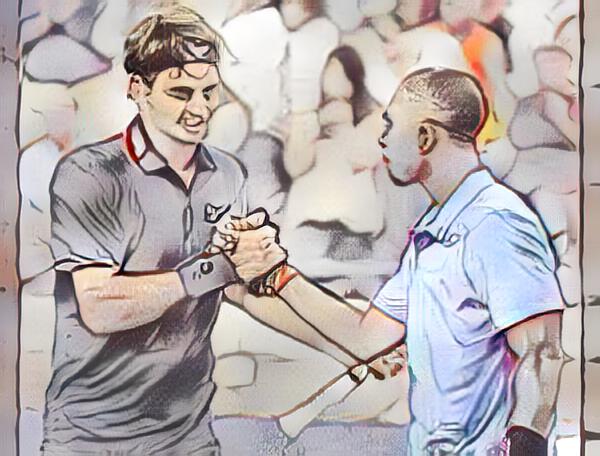}&
\includegraphics[width=1.9cm]{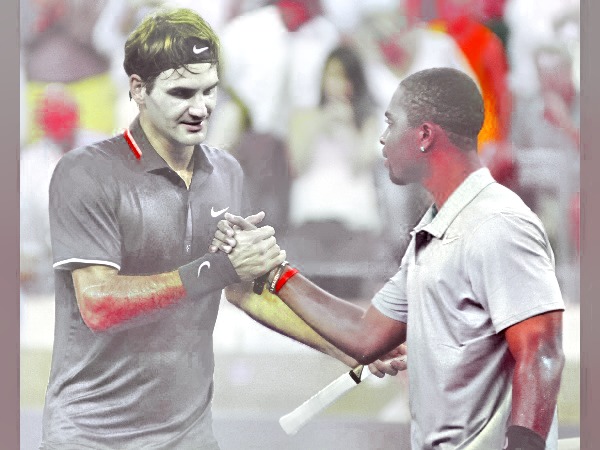}&
\includegraphics[width=1.9cm]{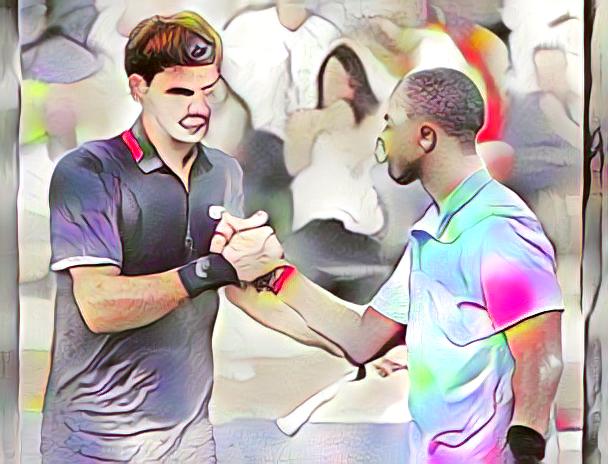}&
\includegraphics[width=1.9cm]{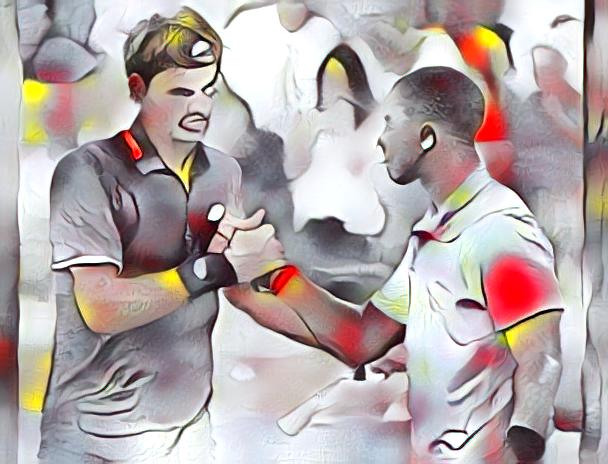}&
\includegraphics[width=1.9cm]{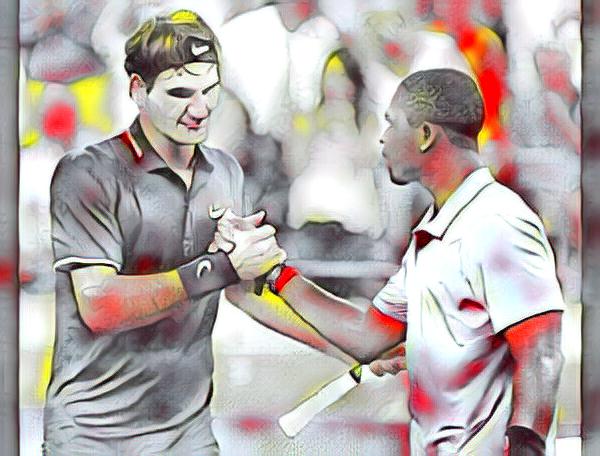}\\

\includegraphics[width=1.9cm]{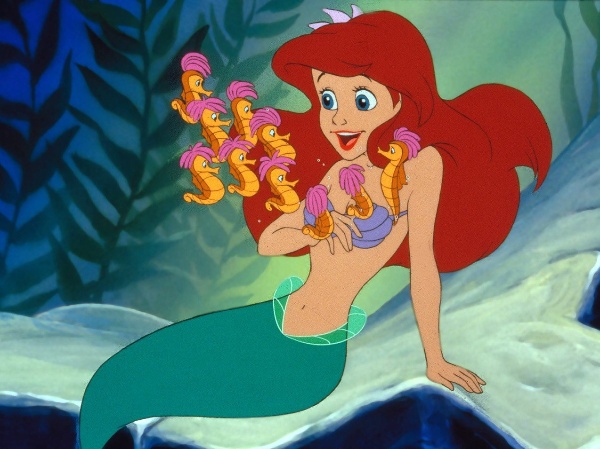}&
\includegraphics[width=1.9cm]{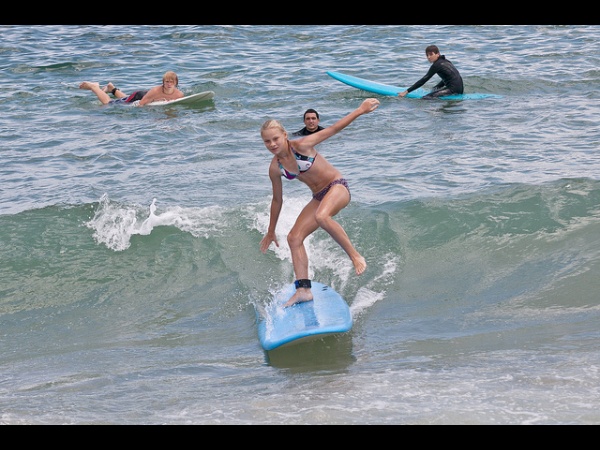}&
\includegraphics[width=1.9cm]{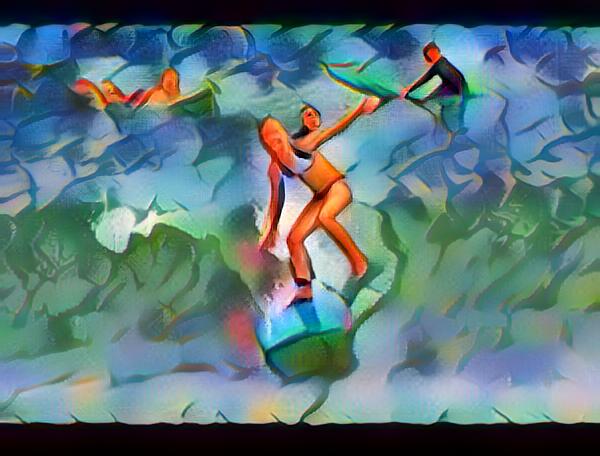}&
\includegraphics[width=1.9cm]{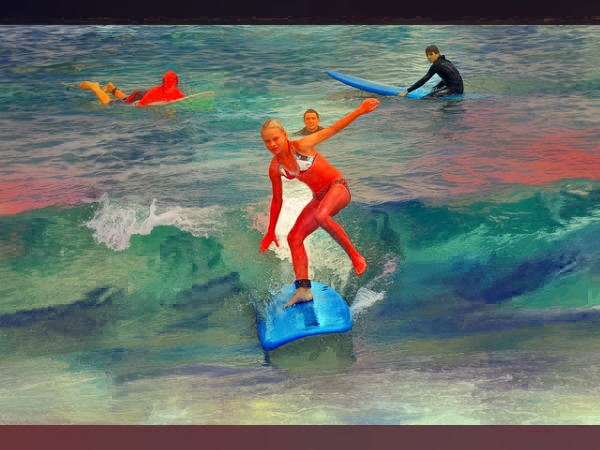}&
\includegraphics[width=1.9cm]{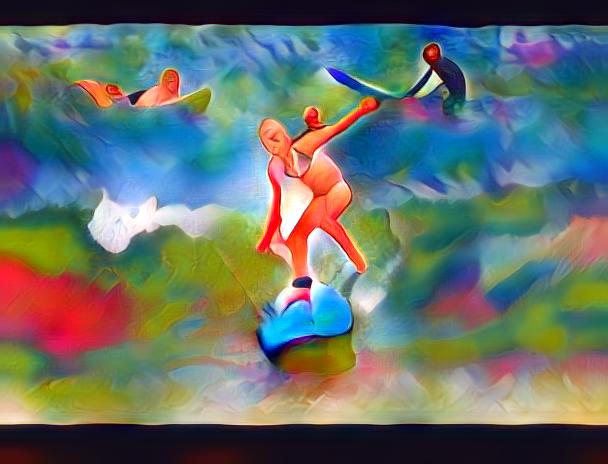}&
\includegraphics[width=1.9cm]{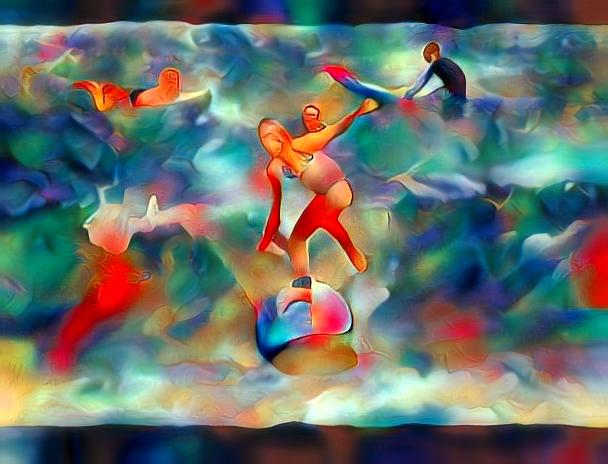}&
\includegraphics[width=1.9cm]{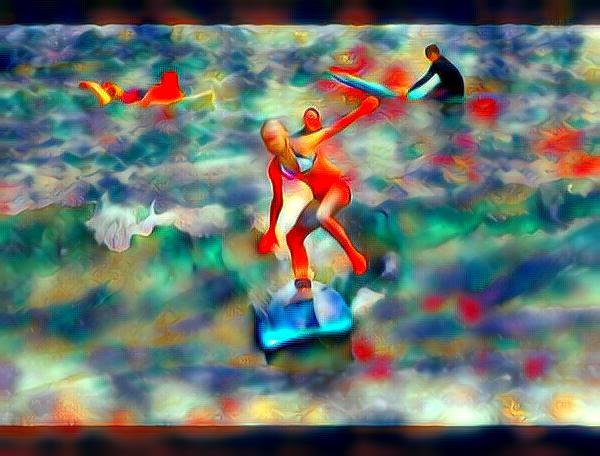}\\
a)&b)&c)&d)&e)&f)&g)\\
\end{tabular}
    \caption{Results from different style transfer methods. Columns: a) style images, b) content images, c) AdaIN, d) PHOTO-R, e) UST-AdaIN, f) UST-WCT, g) UST-WCT4. \label{visual:results}}
\end{figure}

In this section, we compare selected style transfer methods. Fig. \ref{visual:results} shows the example results obtained for different approaches. AdaIN method returns images that are the closest to cartoons from all examined approaches. Unfortunately, it often leaves the content colors or some mixed hues that do not fit the style (especially UST-AdaIN). UST-WCT and UST-WCT4 give results with more appropriate and stylistically coherent colors. However, the results of those models seem to be stylized too much which often leads to significant distortions in those pictures. Moreover, those approaches sometimes struggle with multiple blurs and stains. On the other hand,  PHOTO-R returns very photo-realistic images, it seems that this model transfers almost only color information, which is definitely not what is expected for Comic Style Transfer. Furthermore, one common disadvantage of all methods is that they often carry style and color in an inadequate way, i.e. in Fig. \ref{visual:results} in two last rows there are a lot of red blurs in all result images, while red color should only correspond to the Mickey Mouse shorts and the Little Mermaid's hair.

In addition, we use SSIM index \cite{ssim:} to compare obtained results with corresponding style and content images. Result values are presented in Table \ref{ssim:results}. In terms of content comparison, the biggest value of SSIM index we obtain for PHOTOR-R method which is expected due to its photo-realistic nature. Morover, greater SSIM index values for AdaIN and UST-WCT4 methods compared to UST-WCT and UST-AdaIN are probably caused by less higher-level statistics carried by this approaches (Result images are less distorted in comparison with UST-WCT and UST-AdaIN). Unfortunately, SSIM index does not work well for style comparison. All aproaches gives very similar SSIM values, so we can not conclude wich one performs better. The largest value for PHOTO-R only confirms this thesis, as this method transfers almost only color information without higher-level statistics. This, in turn, is not what we expect for Style Transfer comparison.

\begin{table}[H]
\centering
\bgroup
\def\arraystretch{1.1}
\setlength\tabcolsep{2.6em}
\caption{Average SSIM index values between result images and content/style images for specific approaches. \label{ssim:results}}
\begin{tabular}{| c | c | c |}
\hline
Method&SSIM style&SSIM content\\
\hline
AdaIN&0.135419&0.282729\\
\hline

UST-AdaIN&0.157140&0.214915\\
\hline

UST-WCT&0.146336&0.191519\\
\hline

UST-WCT4&0.134137&0.267988\\
\hline

PHOTO-R&0.161805&0.651781\\
\hline

\end{tabular}

\egroup
\vspace{-8mm}
\end{table}

\subsection{Quantitative Results}

In addition to the qualitative assessment, we also tested the efficiency of each method. In this subsection, we compare the selected approaches in terms of the stylization speed. In Table. \ref{speed:results} we can see the average time of the stylization of one image with a resolution of $600 \times 450$ pixels. The content and style images have the same size. We provide an average runtime for 400 executions on non-repeatable input images. Our tests are performed on a machine with 12GB TITAN X GPU. The third column in Fig. \ref{speed:results} represents the number of styles that every model of each algorithm can produce. All methods except PHOTO-R give very competitive results. Slight differences are connected with complexity of each model. For example AdaIN layer in AdaIN and UST-AdaIN method is computationally faster than WCT in UST-WCT and UST-WCT4 approaches, which is caused by eigenvalue decomposition step in WCT \cite{universal:style}. Only PHOTO-R model is a little slower than others, but definitely fast enough for a real-time use.

\begin{table}[t]
\centering
\bgroup
\def\arraystretch{1.1}
\setlength\tabcolsep{2.6em}
\caption{Average speed comparison of selected Fast Neural Style Transfer algorithms for images of size 600x450 pixels (on 12GB TITAN X GPU).\label{speed:results}}
\begin{tabular}{| c | c | c |}
\hline
Method&Time(s) for 600x450 pixel images&Styles/Model\\
\hline
AdaIN&0.201&$\infty$\\
\hline

UST-AdaIN&0.378&$\infty$\\
\hline

UST-WCT&0.831&$\infty$\\
\hline

UST-WCT4&0.510&$\infty$\\
\hline

PHOTO-R&1.697&$\infty$\\
\hline

\end{tabular}

\egroup
\end{table}

\section{Survey}

To evaluate the selected approaches we conducted a survey on over 100 participants, living in Poland and with the age range between 18 to 70 years with different professions, including students, retired and others. 
It took us two weeks to gather all the votes. We prepared different examples of comic style transfer using five chosen models. In each question we asked which image transfers comic style from a given image to presented photo in the best way. Survey answers were distributed randomly in order to prevent choosing the same answers in every question. The main goal was to obtain subjective opinions of as many various people as possible. The quality measures were ability to preserve content information, style transfer and how "comic" is final image. 

\begin{figure}[H]
  \centering
  \includegraphics[width=1.0\textwidth]{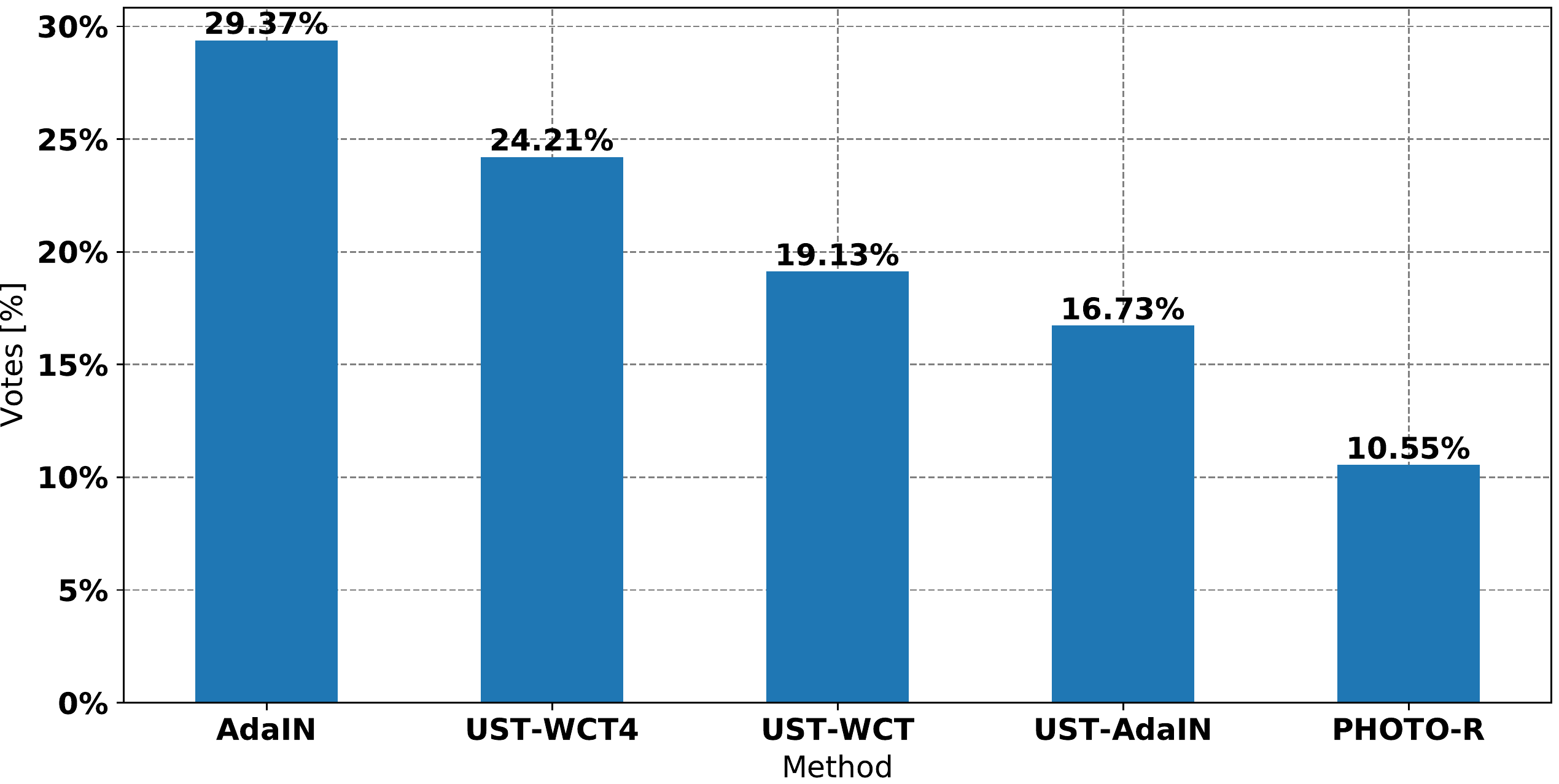}
  \caption{Survey results.\label{survey:results}}
\end{figure}

Survey results are presented in Fig. \ref{survey:results}. As we can see the best results were obtained by the AdaIN approach. It confirms our assumptions that this method gives results that are the closest to cartoon or comics in terms of stylistic similarity. The second and the third place was assigned to UST-WCT4 and UST-WCT, respectively. Better results for UST-WCT4 are probably caused by less higher-level statistics carried by this model with respect to the UST-WCT approach, so the result images are less distorted.

\section{Conclusion}

In this work, we compared five different Neural Style Transfer approaches and how they cope with Transferring Comic Style. Experimental results demonstrate pros and cons of the evaluated approaches. Each of them suffers from similar problems such as blur effects or inappropriate color transfer and we believe that there is still some place for improvement. As future work we intend to introduce Inter-layer Correlations (as described in M. Yeh \textit{et al.} paper \cite{inter:layer}) to all evaluated methods in order to check whether it can improve Comic Style Transfer quality. 

%
%

\end{document}